\documentclass{esannV2}
\usepackage[dvips]{graphicx}
\usepackage[latin1]{inputenc}
\usepackage{amssymb,amsmath,array}

\usepackage{todonotes}
\usepackage{hyperref}
\usepackage{wrapfig}
\usepackage{subcaption}
\usepackage{arydshln} 
\usepackage{booktabs} 
\usepackage{siunitx}  

\begin{document}
\title{Application of Graph Convolutions in a Lightweight Model for Skeletal Human Motion Forecasting}

\author{Luca Hermes, Barbara Hammer and Malte Schilling
\thanks{This research was supported by the research training group ``Dataninja'' (Trustworthy AI for Seamless Problem Solving: Next Generation Intelligence Joins Robust Data Analysis) funded by the German federal state of North Rhine-Westphalia.}
%
%
\vspace{.3cm}\\
%
Machine Learning Group, Bielefeld University \\
33501 Bielefeld, Germany
}

\maketitle

\begin{abstract}
Prediction of movements is essential for successful cooperation with intelligent systems.
We propose a model that integrates organized spatial information as given through the moving body's skeletal structure. This inherent structure is exploited in our model through application of Graph Convolutions and we demonstrate how this allows leveraging the structured spatial information into competitive predictions that are based on a lightweight model that requires a comparatively small number of parameters.
\end{abstract}

\section{Introduction}

Human motion forecasting has many useful applications. As intelligent systems should interact with humans, it becomes necessary to predict human movements and actions. Take, as one example, a cooperative task in which a human and a robot should safely collaborate in an assembly process. Therefore, multiple techniques have been developed to tackle such tasks. Ranging from simple heuristic approaches towards learning-based approaches. Machine learning approaches for such temporal data were mostly relying on recurrent networks that were applied to vectorized representations of joints. These were using skeletal structure only implicitly. The latest advances in geometric deep learning (GDL) provide a direct way to leverage the skeletal structure for predictions via graph convolutions (GCNs). In this work, we propose a straightforward implementation of such a model. We use spatio-temporal convolutions together with a type of GCN to extract spatio-temporal features in human motion, resulting in a simple autoregressive model. We adapt dilated causal convolutions for temporal modeling as used in \cite{wavenet}, but include local joint connectivity which leads to a lightweight spatio-temporal operation. The code will be made publicly available.\footnote{\href{https://github.com/LucaHermes/lightweight-motion-forecasting}{https://github.com/LucaHermes/lightweight-motion-forecasting}}

\section{Methods}
In this section, we introduce our deep geometric model for motion forecasting. The model is based on \textit{Graph-WaveNet} \cite{graph_wavenet}, a spatio-temporal extension to the original \textit{WaveNet} \cite{wavenet}.

\begin{wrapfigure}{r}{0.48\textwidth}
    \centering
    \includegraphics[width=0.465\textwidth]{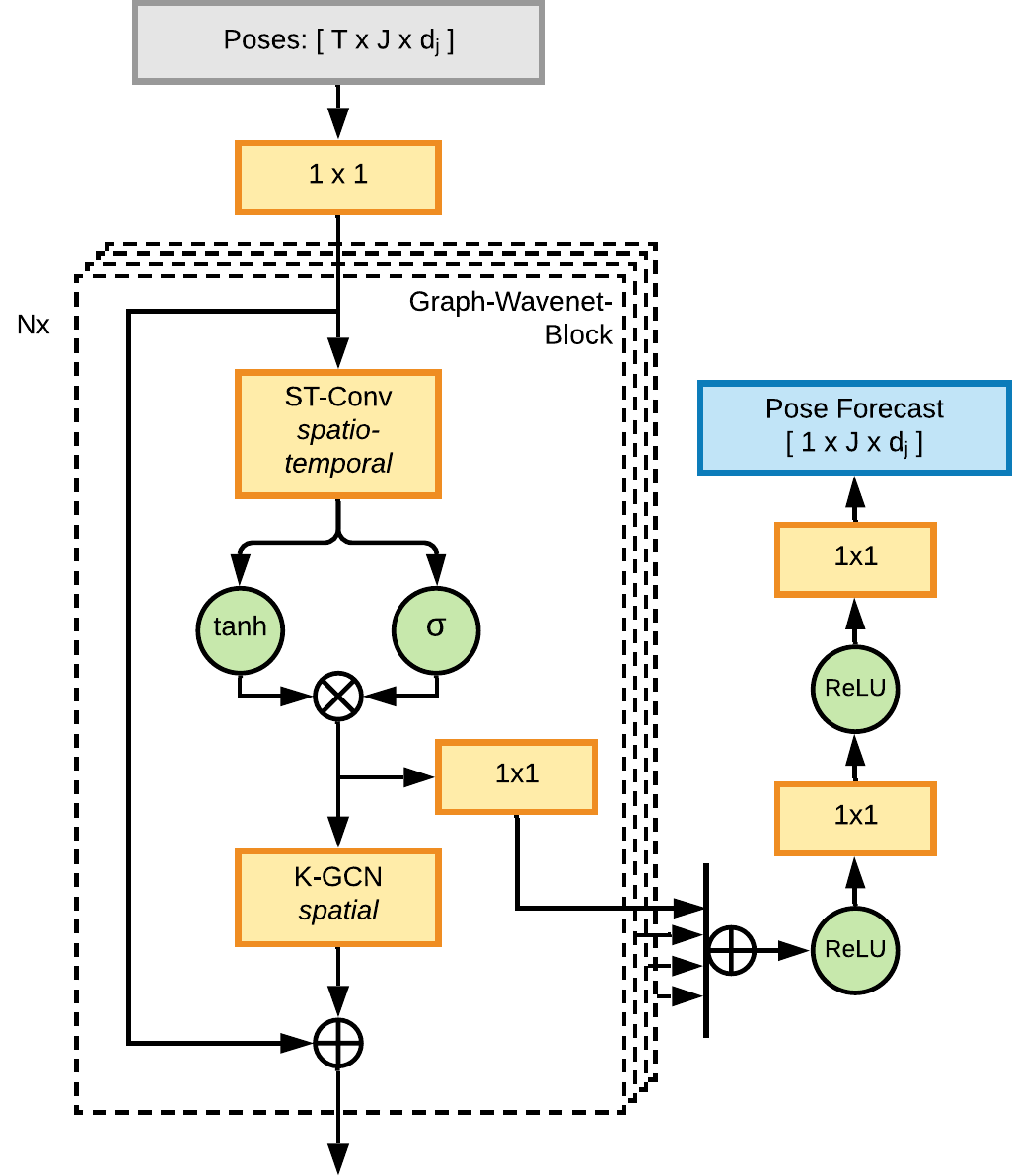}
    \caption{Model architecture, with $N$ consecutive spatio-temporal processing blocks, followed by two alternating ReLU and linear layers ($1 \times 1$). Rounded arrows denote a channel-split.} 
    \label{fig:graph_wavenet_enc}
\end{wrapfigure}

Input to the model is a time series of $T$ consecutive skeletal states with $J$ joints and a $d_j$-dimensional joint representation, $\mathbf{S} \in \mathbb{R}^{T \times J \times d_j}$ (Fig. \ref{fig:graph_wavenet_enc} shows the model architecture). In our experiments, we represent joints as quaternions, i.e. $d_j = 4$.

First, the linear input layer ($1 \times 1$) is applied to each joint and acts as a trainable embedding for the $d_j$-dimensional joint inputs.
Secondly, $N$ consecutive Graph-WaveNet blocks extract spatio-temporal features from the given time series. Every block produces a skip output and the sum of these outputs are, thirdly, passed to an MLP that is applied per joint.

Every Graph-WaveNet block performs a spatio-temporal convolution (ST-Conv) followed by a purely spatial graph-convolution (K-GCN), as shown in Fig. \ref{fig:graph_wavenet_enc}. 
Both operations use the same output dimensionality, to which we refer to as the block dimensionality $d_b$. A residual connection is applied which bypasses both convolutions by adding the block input to the block output. 
The result is fed to the subsequent block. 

We use the following model configuration in our experiments: The input layer consists of $64$ neurons. We use $N = 5$ blocks with $d_b = 64$ and a skip output dimensionality of $256$. The two layers in the output MLP have $256$ and $4$ neurons, respectively. The resulting model has $\SI{4.46e5}{}$ trainable parameters.

\vspace{0.4cm}
\noindent\textbf{Spatio-Temporal Convolution} \hspace{0.5cm} In the original Graph-WaveNet a purely temporal convolution is used to extract temporal features from the input. 
In contrast, we substitute the temporal convolution that acts on each joint individually with a spatio-temporal convolution that integrates information from neighboring joints. This operation is inspired by the temporal extension module (TEM) \cite{temporal_extension_module}, which is a type of GCN placed prior to a temporal operation. Therefore, the temporal operation operates on information from the neighborhood $\mathcal{N}_j$.

We integrate the idea of TEM directly into the spatio-temporal convolution operation.
This means, instead of applying a 1-dimensional convolution on the trajectory of a single joint, we apply a 2-dimensional spatio-temporal convolution on the trajectory of a kinematic chain of joints. 
The joint hierarchy is given by the kinematic tree of the skeleton. Through selecting the hip as the root joint, an ordering is introduced that extends into the leaf-joints, i.e. the head, hands and foot joints.

Fig. \ref{fig:tec_application} visualizes our spatio-temporal convolution. First, the parent and grandparent joint for every joint $j$ are sampled based on the kinematic tree.
Afterwards, the trajectories for these three joints are stacked and convolved using a convolutional kernel $\mathbf{W} \in \mathbb{R}^{\tau_s \times \tau_t \times d_{in} \times d_{out}}$ (blue rectangle). The kernel size is denoted as $\tau_s$ and $\tau_t$ in the spatial and temporal dimension, respectively. The applied convolution is causal and dilated in the temporal dimension following the pattern of \cite{wavenet, graph_wavenet}.

\begin{figure}[h!]
    \captionsetup[subfigure]{position=b}
    \centering
    \subcaptionbox{\label{fig:tec_joint_mapping}}
    {\includegraphics[width=.3\linewidth]{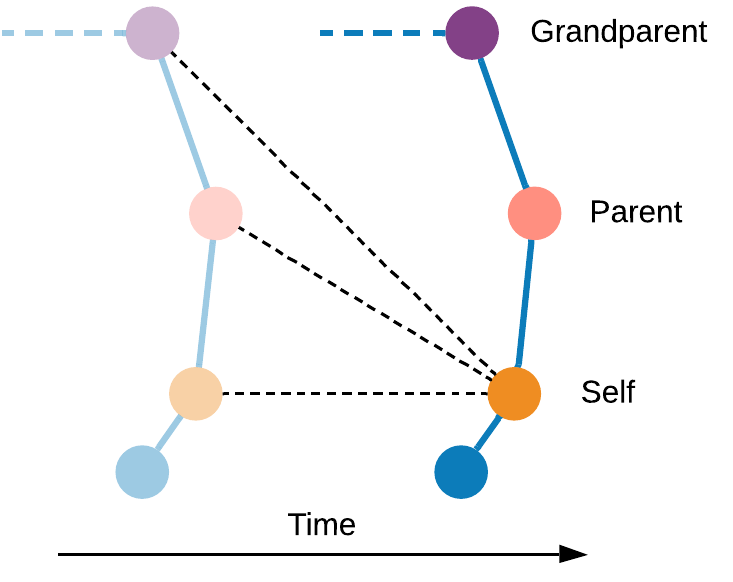}}\hspace{.5cm}
    \subcaptionbox{\label{fig:tec_application}}
    {
    \begin{minipage}[t][][t]{.4\linewidth}
        \centering
        \hfill
        \includegraphics[width=\linewidth]{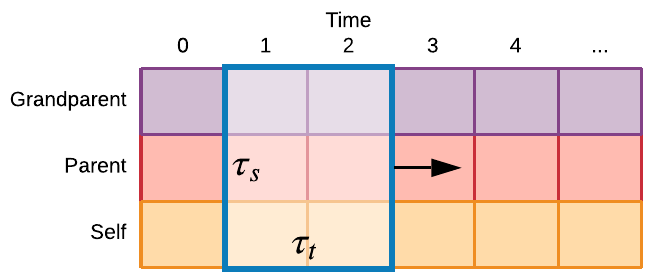}
        \hfill
    \end{minipage}%
    }
    \caption{(a) Samplig strategy for the ST-Conv. (b) Computation of the spatio-temporal convolution. A 2-dimensional kernel computes the convolution of the time-series from three joints. The kernel size in the spatial and the temporal domain is denoted by $\tau_s$ and $\tau_t$, respectively.}
    \label{fig:tec_application}
    \vspace{-0.1cm}
\end{figure}

\vspace{0.4cm}
\noindent\textbf{Spatial Graph-Convolution} \hspace{0.5cm} The purely spatial convolution is computed using graph convolutions (GCNs) as defined in \cite{kipf_welling_17_spacial_gcn}.
Instead of using the full skeletal graph, we again utilize the kinematic tree that was described above to convert the undirected skeletal graph into three directed subgraphs: In the first subgraph, an edge $e=(j, u)$ exists if and only if joint $j$ relative to joint $u$ is further up in the hierarchy of the kinematic tree. This subgraph retains all edges linking a joint to its immediate child joints. The second subgraph is similarly constructed, but with inverted edge direction, therefore retaining all edges linking a joint to its immediate parent joints.
The third subgraph consists only of self-loops.
The subgraphs are represented by three adjacency matrices $\mathbf{A}_i$. Note that $\mathbf{A}_2$ corresponds to the identity matrix. The output of this operation is computed as follows: 

$$
    \text{K-GCN}(\mathbf{X}_{in}) = \sum_{i = 0}^{2}\mathbf{D}_i^{-1}\mathbf{A}_i \mathbf{X}_{in} \mathbf{W}_i
$$

\noindent The inputs are given as $\mathbf{X}_{in}$, $\mathbf{W}_i$ is the parameter matrix of the $i^{th}$ GCN operation and $\mathbf{D}_i$ denotes the degree matrix of subgraph $i$.

\section{Results}
In this section, we describe the experiments we conducted on the presented model. 
To quantify the performance, we perform an evaluation using protocols from the related literature and provide qualitative results.

\vspace{0.4cm}
\noindent\textbf{Dataset and Training Setup} \hspace{0.3cm} 
We train and evaluate the model on the Human3.6M (H3.6M) dataset \cite{h36m_pami, h36m_ionescu}. The dataset consists of motion capture data from seven different human subjects performing 15 different actions. 
Table \ref{tab:quantitative_forecasting_std} and \ref{tab:quantitative_forecasting_stdII} provide an overview of these actions.
We follow the general preprocessing procedure in \cite{quaternet}, 
downsampling the dataset by a factor of 2 to 25 Hz and mirror each trajectory across the $y$-$z$-plane.

We split the data by subject into a train (subjects 1, 7, 8, 9, 11), validation (subject 6) and test set (subject 5).
The models are trained using a batch size of 16 trajectories over 3000 epochs on the mean absolute error in the quaternion space.
An epoch consists of five samples from every sequence. A sequence is a single trial of one subject, performing a single action. 
Each sample contains successive frames from a single downsampled or mirrored version of the trajectories.
The model is trained using the Adam optimizer with an initial learning rate of $\eta = 0.001$ that is decayed exponentially by a factor of $\alpha = 0.999$ after every epoch.
We use a seed trajectory length of 32 frames ($= 1.28 sec$) to condition the model, because this perfectly fits into the receptive field. The target sequence has a length of 10 frames ($= 400 ms$) and was generated autoregressively.

\vspace{0.4cm}
\noindent\textbf{Model Evaluation} \hspace{0.3cm} 
We show qualitative and quantitative results for the proposed model.
For our model evaluation, we use the non-mirrored, but downsampled dataset. 
For the quantitative evaluation, we follow the standard evaluation protocol of \cite{fragkiadaki2015}.
This protocol first constructs an evaluation set that consists of four random samples from every sequence in the test set, i.e. the trials of subject 5.
\footnote{We were able to draw the same sequences as \cite{fragkiadaki2015} using the \textit{RandomState} implementation of the \textit{random} package of \textit{NumPy} (v. 1.19.2) with a seed of $s = 1234567890$.}
An error metric quantifies the performance of the model on this evaluation set.
The metric used by \cite{fragkiadaki2015} is the Euclidean distance between the predicted and target rotations converted to Euler angles. The following equation summarizes this metric:

$$
    d(\mathbf{x}, \mathbf{y}) = \frac{1}{T} \sum_{t=0}^T \sqrt{  \sum_j^J \sum_d (x_{t,j,d} - y_{t,j,d})^2  },
$$

\noindent where $T$ is the number of time-steps, $J$ is the number of joints and the sum over $d$ accumulates the error in the $x$, $y$ and $z$ dimension of the given Euler angles. The final results correspond to the average taken over the four samples.
Following \cite{martinez2017}, the results for a running average over 2 and 4 frames (Run. avg. 2/4) and a zero-velocity-model are also documented as baselines. The zero-velocity model returns the first observed frame as the prediction for all successive frames.

Table \ref{tab:quantitative_forecasting_std} shows the results under the standard protocol for four actions. The running average and zero-velocity baselines are included, as well as results for multiple SoTA models. 
Table \ref{tab:quantitative_forecasting_stdII} lists additional results for the remaining 12 actions (we include results from other models when available for comparison). 
Our forecasting model generally shows competitive results that in some cases, e.g. \textit{eating}, \textit{directions}, and \textit{walkingtogether}, outperform the referenced approaches. The original Graph-WaveNet architecture converges to a zero-velocity model.

\begin{table}[h]
    \centering
    \caption{Quantitative results on action forecasting under the standard protocol of \cite{fragkiadaki2015}. We further specify the input and output mode of some models (\textit{input/output}), where either velocities (\textit{vel.}) or absolute angles (\textit{abs.}) are used.
    }
    \label{tab:quantitative_forecasting_std}
    \resizebox{\textwidth}{!}{
    \tabcolsep=0.11cm
    \begin{tabular}{l llll l llll l llll l llll || l}
    \hline
                              & \multicolumn{4}{l}{Walking} & & \multicolumn{4}{l}{Eating} & & \multicolumn{4}{l}{Smoking} & & \multicolumn{4}{l||}{Discussion} & No. of \\ \cline{2-5}\cline{7-10}\cline{12-15}\cline{17-20}
               Milliseconds   & 80   & 160  & 320  & 400    & & 80   & 160  & 320  & 400   & & 80   & 160  & 320  & 400    & & 80   & 160  & 320  & 400 & Parameters \\ \toprule
        Run.-avg. 4   & 0.64 & 0.87 & 1.07 & 1.20   & & 0.40 & 0.59 & 0.77 & 0.88  & & 0.37 & 0.58 & 1.03 & 1.02   & & 0.60 & 0.90 & 1.11 & 1.15 & - \\ 
        Run.-avg. 2  & 0.48 & 0.74 & 1.02 & 1.17   & & 0.32 & 0.52 & 0.74 & 0.87  & & 0.30 & 0.52 & 0.99 & 0.97   & & 0.41 & 0.74 & 0.99 & 1.09 & - \\ 
        Zero-velocity  & 0.39 & 0.68 & 0.99 & 1.15   & & 0.27 & 0.48 & 0.73 & 0.86  & & 0.26 & 0.48 & 0.97 & 0.95   & & 0.31 & 0.67 & 0.94 & 1.04 & - \\
LSTM-3LR \cite{fragkiadaki2015} & 0.77 & 1.00 & 1.29 & 1.47& & 0.89 & 1.09 & 1.35 & 1.46  & & 1.34 & 1.65 & 2.04 & 2.16   & & 1.88 & 2.12 & 2.25 & 2.23 &  $\SI{1.48e7}{}$ \\
GRU \textit{sup.} \cite{martinez2017}        & 0.28 & 0.49 & 0.72 & 0.81  & & 0.23  & 0.39 & 0.62 & 0.76  & & 0.33 & 0.61 & 1.05 & 1.15 & & 0.31 & 0.68 & 1.01 & 1.09 & $\SI{3.37e6}{}$ \\
QuaterNet GRU \textit{abs./vel.} \cite{quaternet} & 0.21 & \textit{0.34} & \textit{0.56} & 0.62   & & 0.20 & 0.35 & 0.58 & 0.70  & & 0.25 & 0.47 & 0.93 & 0.90   & & 0.26 & 0.60 & 0.85 & \textit{0.93} & $\SI{9.5e6}{}$ \\
QuaterNet CNN \textit{abs./vel.} \cite{quaternet} & 0.25 & 0.40 & 0.62 & 0.70   & & 0.22 & 0.36 & 0.58 & 0.71  & & 0.26 & 0.49 & 0.94 & 0.90   & & 0.30 & 0.66 & 0.93 & 1.00 & $\SI{8.8e6}{}$ \\
DCT-GCN \textit{short-term} \cite{dct_gcn}  & \textbf{0.18} & \textbf{0.31} & \textbf{0.49} & \textbf{0.56} & & \textbf{0.16} & \textbf{0.29} & \textit{0.50} & 0.62 & & \textit{0.22} & \textit{0.41} & 0.86 & \textit{0.80}   & & \textbf{0.20} & \textbf{0.51} & \textbf{0.77} &  \textbf{0.85} & $\SI{2.6e6}{}$ \\
DMGNN \cite{li_multiscaleGCN2020}  & \textbf{0.18} & \textbf{0.31} & \textbf{0.49} & \textit{0.58} & & \textit{0.17} & \textit{0.30} & \textbf{0.49} & \textbf{0.59} & & \textbf{0.21} & \textbf{0.39} & \textbf{0.81} & \textbf{0.77} & & 0.26 & 0.65 & 0.92 & 0.99 & $\SI{6.26e7}{}$ \\ \hdashline
    Ours \textit{abs./vel.} & 0.23 & 0.37 & 0.61 & 0.69 & & 0.18 & 0.31 & 0.54 & 0.66 & & 0.23 & 0.46 & 0.93 & 0.90 & & 0.31 & 0.70 & 0.97 & 1.07 & $\SI{4.46e5}{}$ \\

    Ours \textit{vel./vel.} & \textit{0.19} & \textit{0.34} & 0.57 & 0.63 & & \textbf{0.16} & \textbf{0.29} & \textit{0.50} & \textit{0.60} & & \textit{0.22} & \textit{0.41} & \textit{0.85} & 0.81 & & \textit{0.22} & \textit{0.57} & \textit{0.84} & \textit{0.93} & $\SI{4.46e5}{}$ \\ 

    \hline
    \end{tabular}
    }
\end{table}

\begin{table}[h]
    \centering
    \caption{Quantitative results for the remaining actions; Continuation of table \ref{tab:quantitative_forecasting_std}.}
    \label{tab:quantitative_forecasting_stdII}
    \resizebox{\textwidth}{!}{
    \tabcolsep=0.11cm
    \begin{tabular}{l llll l llll l llll l llll}
    \hline
                              & \multicolumn{4}{l}{Directions} & & \multicolumn{4}{l}{Greeting} & & \multicolumn{4}{l}{Phoning} & & \multicolumn{4}{l}{Posing} \\ \cline{2-5}\cline{7-10}\cline{12-15}\cline{17-20}
               Milliseconds   & 80   & 160  & 320  & 400    & & 80   & 160  & 320  & 400   & & 80   & 160  & 320  & 400    & & 80   & 160  & 320  & 400  \\ \toprule
    GRU \textit{sup.}         & 0.26 & 0.47 & 0.72 & 0.84   & & 0.75 & 1.17 & 1.74 & 1.83   & & \textbf{0.23} & \textbf{0.43} & \textbf{0.69} & \textbf{0.82}    & & 0.36 & 0.71 & 1.22 & 1.48 \\
    DCT-GCN \textit{st}       & 0.26 & 0.45 & 0.71 & \textit{0.79}   & & \textit{0.36} & \textbf{0.60} & \textit{0.95} & \textit{1.13}   & & 0.53 & 1.02 & 1.35 & 1.48    & & \textbf{0.19} & \textbf{0.44} & \textbf{1.01} & \textbf{1.24} \\
    DMGNN                     & \textit{0.25} & \textit{0.44} & \textbf{0.65} & \textbf{0.71}   & & \textit{0.36} & \textit{0.61} & \textbf{0.94} & \textbf{1.12}   & & \textit{0.52} & \textit{0.97} & 1.29 & 1.43     & & \textit{0.20} & \textit{0.46} & 1.06 & 1.34 \\
               \hdashline
    Ours \textit{abs./vel.}   & 0.32 & 0.47 & \textit{0.68} & 0.80 & & 0.42 & 0.72 & 1.14 & 1.36 & & 0.54 & 1.00 & 1.34 & 1.47 & & 0.27 & 0.55 & \textit{1.05} & \textit{1.27} \\

    Ours \textit{vel./vel.}   & \textbf{0.24} & \textbf{0.43} & 0.77 & 0.81 & & \textbf{0.35} & \textit{0.61} & 1.01 & 1.20 & & 0.53 & 1.00 & \textit{1.28} & \textit{1.40} & & 0.26 & 0.51 & 1.08 & 1.32 \\ 

    \hline
                              & \multicolumn{4}{l}{Purchases} & & \multicolumn{4}{l}{Sitting} & & \multicolumn{4}{l}{Sittingdown} & & \multicolumn{4}{l}{Takingphoto} \\ \cline{2-5}\cline{7-10}\cline{12-15}\cline{17-20}
               Milliseconds   & 80   & 160  & 320  & 400    & & 80   & 160  & 320  & 400   & & 80   & 160  & 320  & 400    & & 80   & 160  & 320  & 400  \\ \toprule
    GRU \textit{sup.}         & 0.51 & 0.97 & 1.07 & 1.16   & & 0.41 & 1.05 & 1.49 & 1.63   & & 0.39 & 0.81 & 1.40 & 1.62   & & 0.24 & 0.51 & 0.90 & 1.05 \\
    DCT-GCN \textit{st}       & 0.43 & \textit{0.65} & \textit{1.05} & \textbf{1.13}   & & \textit{0.29} & \textit{0.45} & \textit{0.80} & \textbf{0.97}   & & \textit{0.30} & \textbf{0.61} & \textbf{0.90} & \textbf{1.00}   & & \textbf{0.14} & \textbf{0.34} & \textbf{0.58} & \textbf{0.70} \\
    DMGNN                     & \textbf{0.41} & \textbf{0.61} & \textit{1.05} & \textit{1.14}   & & \textbf{0.26} & \textbf{0.42} & \textbf{0.76} & \textbf{0.97}   & & 0.32 & \textit{0.65} & \textit{0.93} & \textit{1.05}   & & \textit{0.15} & \textbf{0.34} & \textbf{0.58} & \textit{0.71} \\
\hdashline
    Ours \textit{abs./vel.}  & 0.56 & 0.75 & \textbf{1.03} & 1.15 & & 0.31 & 0.50 & 0.91 & 1.12 & & 0.33 & \textit{0.65} & 0.96 & 1.09 & & 0.19 & \textit{0.42} & 0.73 & 0.93 \\ 

     Ours \textit{vel./vel.} & \textit{0.42} & \textbf{0.61} & 1.08 & 1.15 & & 0.30 & 0.49 & 0.90 & \textit{1.09} & & \textbf{0.29} & \textit{0.65} & 0.97 & 1.08 & & \textit{0.15} & \textbf{0.34} & \textbf{0.58} & 0.72 \\ 

    \hline
                              & \multicolumn{4}{l}{Waiting} & & \multicolumn{4}{l}{Walkingdog} & & \multicolumn{4}{l}{Walkingtogether} & & \multicolumn{4}{l}{\textit{Average}} \\ \cline{2-5}\cline{7-10}\cline{12-15}\cline{17-20}
               Milliseconds   & 80   & 160  & 320  & 400    & & 80   & 160  & 320  & 400   & & 80   & 160  & 320  & 400    & & 80   & 160  & 320  & 400  \\ \toprule
    GRU \textit{sup.} $\quad\quad\quad$   & 0.28 & 0.53 & 1.02 & \textit{1.14}   & & 0.56 & 0.91 & 1.26 & 1.40    & & 0.31 & 0.58 & 0.87 & 0.91   & & 0.36 & 0.67 & 1.02 & 1.15 \\
    DCT-GCN \textit{st}  & \textit{0.23} & \textit{0.50} & \textit{0.91} & \textit{1.14}   & & 0.46 & 0.79 & \textit{1.12} & \textit{1.29}    & & \textbf{0.15} & 0.34 & 0.52 & \textit{0.57}   & & \textbf{0.27} & \textbf{0.51} & \textbf{0.83} & \textbf{0.95} \\
    DMGNN                & 0.22 & \textbf{0.49} & \textbf{0.88} & \textbf{1.10}   & & \textbf{0.42} & \textbf{0.72} & 1.16 & 1.34    & & \textbf{0.15} & \textit{0.33} & \textbf{0.50} & \textit{0.57}  & & \textbf{0.27} & \textit{0.52} & \textbf{0.83} & \textbf{0.95} \\
 \hdashline
    Ours \textit{abs./vel.} & 0.25 & 0.51 & 0.93 & \textbf{1.10} & & 0.46 & 0.79 & 1.16 & 1.32 & & \textit{0.17} & 0.37 & 0.58 & 0.65 & & \textit{0.32} & 0.57 & 0.90 & 1.04 \\ 


    Ours \textit{vel./vel.} & \textbf{0.21} & 0.51 & 0.97 & 1.17 & & \textit{0.43} & \textit{0.78} & \textbf{1.10} & \textbf{1.24} & & \textbf{0.15} & \textbf{0.32} & \textbf{0.50} & \textbf{0.54} & & \textbf{0.27} & \textit{0.52} & \textit{0.87} & \textit{0.98} \\ 

    \hline
    \end{tabular}
    }
\end{table}
\begin{figure}
    \centering
    \includegraphics[width=\textwidth]{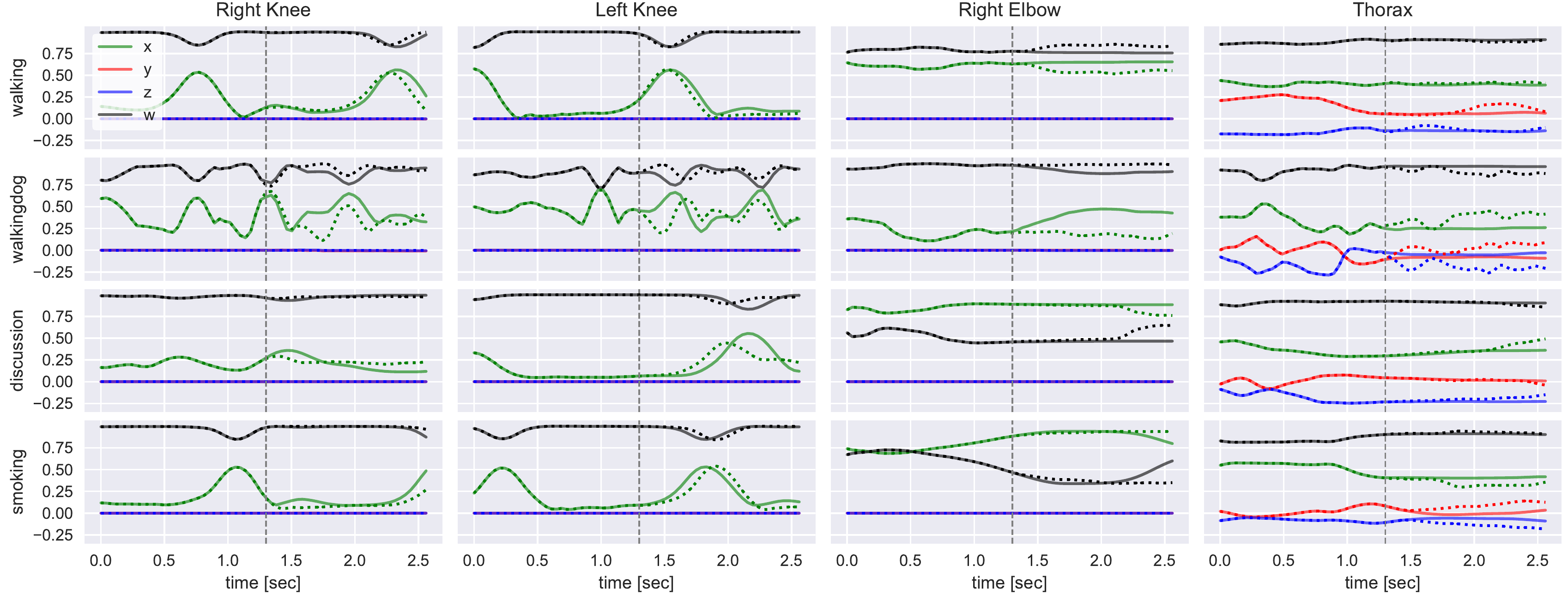}
    \caption{Ground truth (dotted lines) and prediction (solid lines) for four different joints (columns) and four different actions (rows) of the second trial from subject 5. 
    Each plot shows 32 seed frames ($= 1.28 sec$) and 32 target frames.}
    \label{fig:res_joint_level_qualitative_quaternions}
    \vspace{-0.05cm}
\end{figure}

\noindent Fig. \ref{fig:res_joint_level_qualitative_quaternions} visualizes predicted and true individual quaternion dimensions. Overall, the predictions are very similar to ground truth, but finer details are smoothed out. 
This is clearly visible in the trajectory of the right knee joint performing the \textit{walkingdog} action. However, the model is still able to correctly predict the phase timing and thus model major movement features even in the long-term future ($> 400 ms$).

\section{Conclusions}
\label{sec:conclusion}
We combine the well-established causal dilated convolutions from WaveNet with geometric deep learning principles resulting in a lightweight autoregressive model. With about $\SI{4.46e5}{}$ parameters, our model is an order of magnitude less complex compared to current models (cf. Table \ref{tab:quantitative_forecasting_std}).
Nonetheless, it shows competitive results compared to current approaches when evaluated on the H3.6M dataset for skeletal human motion forecasting. 
Furthermore, we show that this model is able to predict the phase timing even in a long-term forecasting setup. 


\begin{footnotesize}

\bibliographystyle{unsrt}
\bibliography{references}

\begin{thebibliography}{10}

\bibitem{wavenet}
A.~van~den Oord, S.~Dieleman, H.~Zen, K.~Simonyan, O.~Vinyals, A.~Graves,
  N.~Kalchbrenner, A.~W. Senior, and K.~Kavukcuoglu.
\newblock Wavenet: {A} generative model for raw audio.
\newblock {\em CoRR}, abs/1609.03499, 2016.

\bibitem{graph_wavenet}
Z.~Wu, S.~Pan, G.~Long, J.~Jiang, and C.~Zhang.
\newblock Graph wavenet for deep spatial-temporal graph modeling.
\newblock {\em CoRR}, abs/1906.00121, 2019.

\bibitem{temporal_extension_module}
Y.~Obinata and T.~Yamamoto.
\newblock Temporal extension module for skeleton-based action recognition,
  2020.

\bibitem{kipf_welling_17_spacial_gcn}
T.~N. Kipf and M.~Welling.
\newblock Semi-supervised classification with graph convolutional networks.
\newblock {\em CoRR}, abs/1609.02907, 2016.

\bibitem{h36m_pami}
C.~Ionescu, D.~Papava, V.~Olaru, and C.~Sminchisescu.
\newblock Human3.6m: Large scale datasets and predictive methods for 3d human
  sensing in natural environments.
\newblock {\em IEEE Transactions on Pattern Analysis and Machine Intelligence},
  36(7):1325--1339, 2014.

\bibitem{h36m_ionescu}
C.~Ionescu, F.~Li, and C.~Sminchisescu.
\newblock Latent structured models for human pose estimation.
\newblock In {\em International Conference on Computer Vision}, 2011.

\bibitem{quaternet}
D.~Pavllo, C.~Feichtenhofer, M.~Auli, and D.~Grangier.
\newblock Modeling human motion with quaternion-based neural networks.
\newblock {\em CoRR}, abs/1901.07677, 2019.

\bibitem{fragkiadaki2015}
K.~Fragkiadaki, S.~Levine, and J.~Malik.
\newblock Recurrent network models for kinematic tracking.
\newblock {\em CoRR}, abs/1508.00271, 2015.

\bibitem{martinez2017}
Julieta Martinez, Michael~J. Black, and Javier Romero.
\newblock On human motion prediction using recurrent neural networks.
\newblock {\em CoRR}, abs/1705.02445, 2017.

\bibitem{dct_gcn}
W.~Mao, M.~Liu, M.~Salzmann, and H.~Li.
\newblock Learning trajectory dependencies for human motion prediction.
\newblock {\em CoRR}, abs/1908.05436, 2019.

\bibitem{li_multiscaleGCN2020}
M.~Li, S.~Chen, Y.~Zhao, Y.~Zhang, Y.~Wang, and Q.~Tian.
\newblock Dynamic multiscale graph neural networks for 3d skeleton-based human
  motion prediction.
\newblock {\em arXiv:2003.08802 [cs.CV]}, 2020.

\end{thebibliography}

\end{footnotesize}


\end{document}